\documentclass{svproc}  
\usepackage{url}
\usepackage{hyperref}
\usepackage{graphicx}
\usepackage{multicol}
\usepackage{footmisc}
\usepackage{booktabs}
\usepackage{float}
\floatstyle{plaintop}
\restylefloat{table}
\usepackage{tabularx}
\usepackage{float}
\usepackage{fancyvrb}
\floatstyle{ruled}
\newfloat{program}{thp}{lop}
\floatname{program}{Program}

\begin{document}
\mainmatter              
\title{
A Framework for Robot Programming in Cobotic Environments: First user experiments}
\titlerunning{Robot Programming by Demonstration}  
%
\author{Ying-Siu Liang*, Damien Pellier*,
Humbert Fiorino*, \and Sylvie Pesty*}
\authorrunning{Ying Siu Liang et al.} 
%
\tocauthor{Y. S. Liang, D. Pellier, and H. Fiorino}
\institute{*Universit\'{e} Grenoble Alpes, CNRS, LIG, F-38000 Grenoble, France\\
\email{firstname.lastname@imag.fr}
}

\maketitle              


\begin{abstract}
The increasing presence of robots in industries has not gone unnoticed. Large industrial players have incorporated them into their production lines, but smaller companies hesitate due to high initial costs and the lack of programming expertise. In this work we introduce a framework that combines two disciplines, Programming by Demonstration and Automated Planning, to allow users without any programming knowledge to program a robot. The user teaches the robot atomic actions together with their semantic meaning and represents them in terms of preconditions and effects. Using these atomic actions the robot can generate action sequences autonomously to reach any goal given by the user. We evaluated the usability of our framework in terms of user experiments with a Baxter Research Robot and showed that it is well-adapted to users without any programming experience.
\end{abstract}

\section{Introduction}
The use of robots has increased productivity and replaced humans for repetitive and manual tasks. However, there remain many tasks that cannot be completely taken over by robots and that still need human intervention, such as high-precision tasks. The introduction of collaborative robots or ``cobots" \cite{colgate1996cobots} opens the door to a safe and barrier-free collaboration between humans and robots. Designed to respond to actions of the human operator, cobots generally perform tasks which humans cannot perform on their own, such as the manipulation of heavy parts. Cobotic systems have been adapted in several industries from the food-processing industry, to aeronautics, to the health industry. Despite the uprising popularity of cobots, there are still many companies which are hesitant with their adoption. In particular, small companies consider the investment cost-ineffective due to high initial costs and, more importantly, the lack of trained personnel who have the required expertise to fully exploit the robots.

Robot Programming by Demonstration (PbD) \cite{billard2008robot} addresses this issue by allowing non-expert users to teach robots new skills by demonstrating a task and without the need for writing machine commands. It is a quick and intuitive programming approach independent of the robot platform. It is an iterative process with the goal to refine the robot's performance by providing repetitive demonstrations. However, existing PbD implementations are not goal-oriented. The robot is generally taught an ordered action sequence to achieve a goal (e.g. stacking or ordering objects on a table \cite{EricM.Orendt2016}), but it cannot deduce an action sequence from a given goal. Teaching full action sequences can be complicated and time-consuming, as the robot has to be taught a new action sequence when the goal changes. Why not simply teach the robot atomic actions that it can use in any arbitrary order and generate the action sequence using an automated planner?

In this work we explore the possibility for a robot to learn action semantics and to generate action sequences autonomously using automated planning techniques. We aim to equip the robot with all the atomic actions needed to act autonomously in any state of the world. We propose a framework where (1) the non-expert user teaches the robot atomic actions by demonstration and builds action models used in automated planning techniques (2) the robot can autonomously generate solutions to any goals specified by the user.
As an initial step we want to investigate the framework's potential usability in terms of user experiments, focusing on the user's ability to understand the logical representations for action models in terms of preconditions and effects. In particular, we want to find out if users with diverse educational backgrounds understand the concept of PbD in relation with these action models and if they find them intuitive. Finally, we are interested in the user's perception of the goal-oriented programming approach.

The remaining of this paper is organised as follows. We will first give an overview of the related work, then we will present our proposed framework. This is followed by the experimental setup, used methods, and experimental results. Finally, we conclude by discussing potential future work.

\section{Related Work}          
\label{S:RelatedWorks}           
Programming by Demonstration (PbD) provides an intuitive medium to allow non-experts, who do not possess the necessary domain knowledge, to communicate skills to robots more easily. The underlying concept is to learn a new skill, also known as a policy, from a set of correct demonstrations provided by the teacher. Research in PbD generally concentrates on learning a good policy in as few demonstrations as possible. Argall et al. \cite{argall2009survey} present a comprehensive overview of different policy derivation techniques, splitting them into three main categories. Mapping functions to approximate a state-to-action mapping for the demonstrated behaviour and often used in combination with other classifiers (e.g. Hidden Markov Models \cite{hovland1996skill}, Neural Networks \cite{pomerleau1991efficient}, k-Nearest Neighbours \cite{saunders2006teaching}); System models use a state transition model in combination with a reward function to learn a policy. With the goal to maximise the cumulative reward over time, the reward function can either be user-defined \cite{smart2002effective} or learned from demonstration data \cite{abbeel2004apprenticeship,atkeson1997robot}.

Our work uses the policy derivation technique of plans, where the policy is represented as a sequence of actions leading from an initial state to a final goal state. Actions are defined in terms of preconditions, i.e. a state of the world that must be attained in order to execute the action, and effects, i.e. the expected state resulting from the action execution. Existing implementations of plans learn action sequences to achieve a single goal starting from different initial states \cite{kuniyoshi1994learning}, but do not learn different action models that can be applied independently. Our approach teaches the robot independent action models that can be used by automated planning techniques to reach any given goal.

Automated planning computationally studies the deliberation process of choosing and organising actions in order to achieve a goal. It can be used to model a robot's skills and strategies, when operating in diverse environments, without the need for expensive hand-coding \cite{ghallab2004automated}.
The focus lies within the development of domain-independent planning systems, namely \textit{planners}, which consist of search algorithms that are not problem-specific. A \textit{planning problem} is given by a set of atomic actions, a description of the state of the world, and some goal state. The planner generates a solution to this problem as an ordered sequence of actions, which guarantees the transition from the initial state to the goal state. Similar to the policy derivation technique of plans in PbD, actions are defined in terms of {preconditions} and {effects}, representing states attained before and after the action execution. Classical planning algorithms use the Planning Domain Definition Language (PDDL)\cite{ghallab2004automated} as their standard encoding language to represent action models. Program \ref{fig:moveObject} shows an example of an action in PDDL ({\tt moveObject(?obj, ?pos1, ?pos2)}), which moves an object ({\tt ?obj}) from one position ({\tt ?pos1}) to another position ({\tt ?pos2}). \vspace{-0.5cm}
  \begin{program}
    \begin{verbatim}
(:action moveObject
    :parameters (?obj - object ?pos1 - position ?pos2 - position)
    :precondition  (and  (at ?obj ?pos1)
                         not(empty ?pos1)
                         (empty ?pos2))
    :effect (and (at ?obj ?pos2)
                 (empty ?pos1)
                 (not (empty ?pos2))))    \end{verbatim}
    \caption{Representation of a moveObject action in PDDL.}
    \label{fig:moveObject}
\end{program}

\noindent
Veeraraghavan et al. \cite{veeraraghavan2008teaching} represent a set of pre-programmed action models in PDDL to learn a sequential task plan from demonstrations. Zita Haigh et al. \cite{ZitaHaigh1998} implement a planning algorithm in a robot but do not teach actions by demonstration.
Our work uses PbD techniques to provide non-expert users the flexibility to teach the robot action models in PDDL, and that are then used by automated planners to generate task plans. As far as we know, there are no user studies in this area which deal with the user's understanding of the logical representations of action models in terms of preconditions and effects.

\section{A Framework for Robot Programming by Demonstration}          
\label{S:Framework}           

\begin{figure}[h]
    \centering
    \includegraphics[scale=0.7]{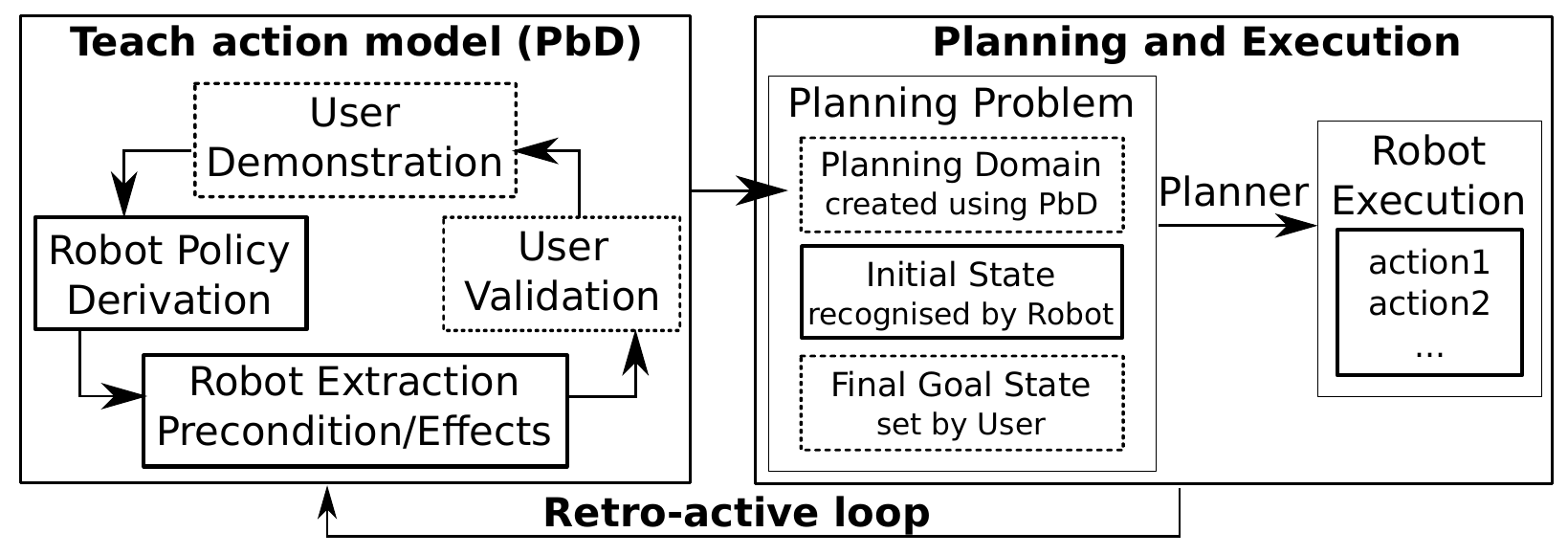}
    \caption{Framework for Robot Programming by Demonstration which use the action models as part of a planning domain and an automated planner to solve a planning problem (dotted lines indicate user actions, solid lines indicate robot actions).}
    \label{fig:framework}
  \end{figure}

Our proposed framework consists of the following four steps: (i) user teaches the robot atomic actions; (ii) robot plans the solution for a defined planning problem; (iii) robot executes the action plan; and (iv) user revisits the learned action models to allow incremental learning. Figure \ref{fig:framework} shows the layout of our proposed framework.  \vspace{-0.5cm}

\vspace{-0.5cm}
\subsection{Teaching Action Models}
The user first needs to construct a planning domain consisting of all atomic actions needed to achieve a goal in that domain.
An action model is created by the user who provides a demonstration of a specific action, e.g. moving a red object from an initial position to a final position ({\tt moveObject(redObj,initPos,finPos)}). Executing an action results in a change in the state of the world such as the changed position of the moved object. By observing these changes before and after the demonstration, the robot extracts preconditions (e.g. red object on initial position) and effects (e.g. red object on final position) to build a generalised action model, known as an \textit{operator}. Table \ref{tab:action-model} compares the states from the demonstrated action to the generalised operator. The generalised operator is automatically translated into PDDL (see Program \ref{fig:moveObject}), allowing the creation of a PDDL planning domain, without the need for any programming knowledge. The user can validate the created operator, modify the proposed preconditions or effects, or provide another demonstration to refine model.
This process is repeated for all atomic actions that are required to build the knowledge base for the complex task.

\begin{table}[h]
\begin{center}
\renewcommand{\arraystretch}{1.3}
\begin{tabularx}{\textwidth}{*{2}{X|}l*{2}{X}}
\toprule
 \textbf{{\tt moveObject}} \hspace{0.02cm}& \textbf{Demonstrated action} & \textbf{Generalised operator} \hspace{0.1cm}\\
 \toprule
{\tt :precondition} \hspace{0.02cm}& \hspace{0.05cm} {\tt (at redObj initPos)} & \hspace{0.05cm} {\tt(at ?obj ?pos1)}\\
 & \hspace{0.05cm} {\tt not(empty initPos)} & \hspace{0.05cm} {\tt not(empty ?pos1)}\\
 & \hspace{0.05cm} {\tt (empty finPos)} & \hspace{0.05cm} {\tt(empty ?pos2)}\\ \hline
{\tt :effect}\hspace{0.02cm} & \hspace{0.05cm} {\tt (at redObj finPos)} & \hspace{0.05cm} {\tt(at ?obj ?pos1)}\\
 & \hspace{0.05cm} {\tt (empty initPos)} & \hspace{0.05cm} {\tt (empty ?pos1)}\\
 & \hspace{0.05cm} {\tt not(empty finPos)} & \hspace{0.05cm} {\tt not(empty ?pos2)}\\
 \hline
 \end{tabularx}
\caption{Generalisation from a demonstrated action to a generalised operator.}\label{tab:action-model}
\end{center}
\end{table}

\subsection{Planning and Execution}
The created planning domain is fed into an automated planner to create a planning problem. The user can specify any goal state that can be reached using the taught actions. The initial state is automatically recognised by the robot by observing the current state of the world. The automated planner generates a plan consisting of an ordered action sequence for the robot to execute. If the user changes the goal, a new plan can be generated accordingly. Generating a plan under different initial states allows the user to test the created operators.
The execution of this action sequence provides the user with the opportunity to test the soundness of the created operators.

\subsection{Retro-active Loop for Incremental Learning}
The execution to a new context is an important step to test the created action models in order to refine them. It is likely that the execution of the plan does not produce the desired outcome, especially if the plan is executed in a context different to the demonstration (e.g. using objects of different colour or shape). Missing preconditions or effects for an action model can lead to suboptimal or non-existing solutions and will need to be corrected by the user.


\section{Experimentation}          
\label{S:Experimentation}           
\begin{figure}[h]
\centering
\begin{minipage}[b]{0.4\textwidth}
  \centering
    \includegraphics[scale=0.2]{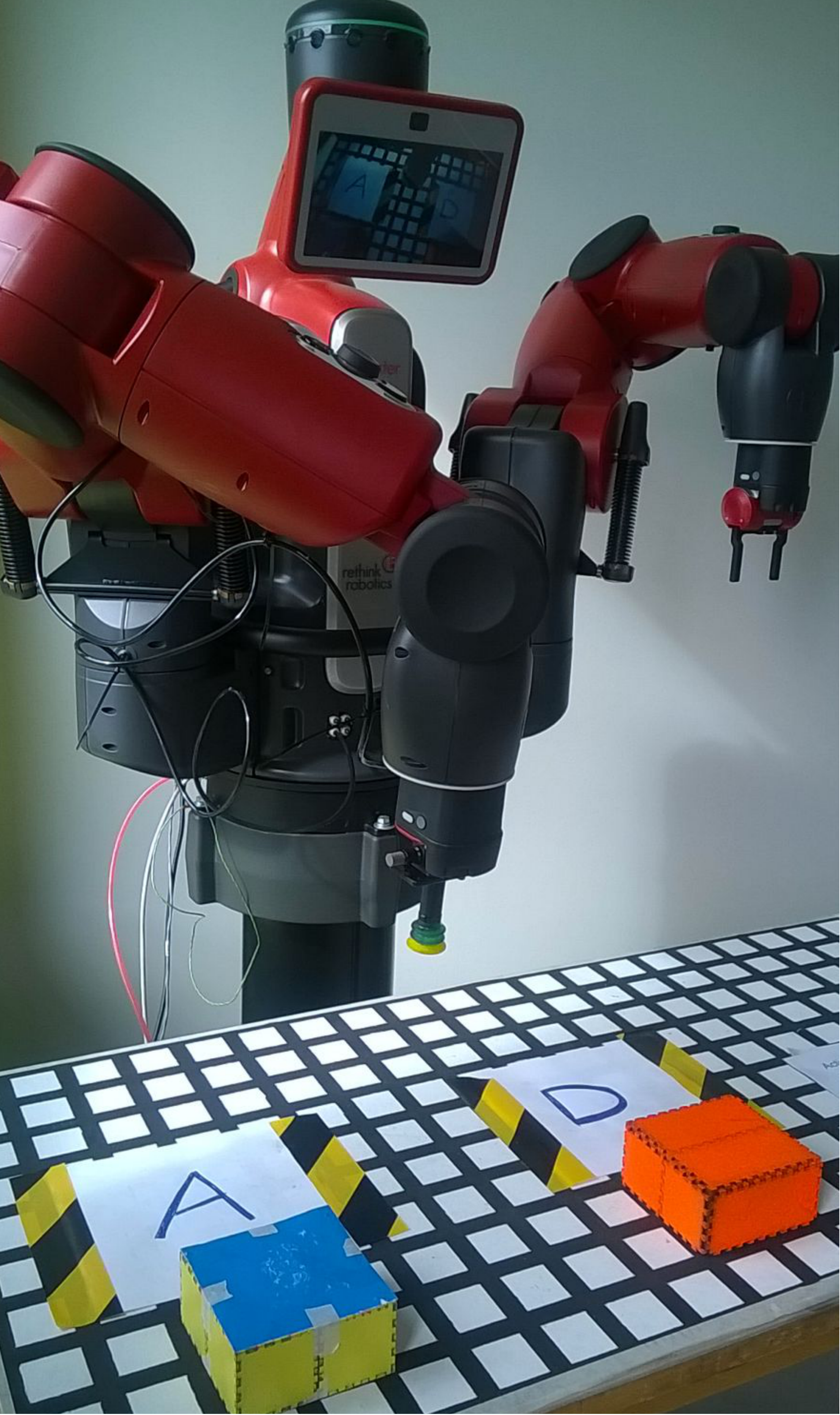}
    \caption{Experimental set up with a Baxter robot and two blocks (blue and red) and positions A (arrival) and D (departure).}
    \label{fig:Experimental set up}
\end{minipage}%
\begin{minipage}[b]{.58\textwidth}
  \centering
    \includegraphics[scale=0.33]{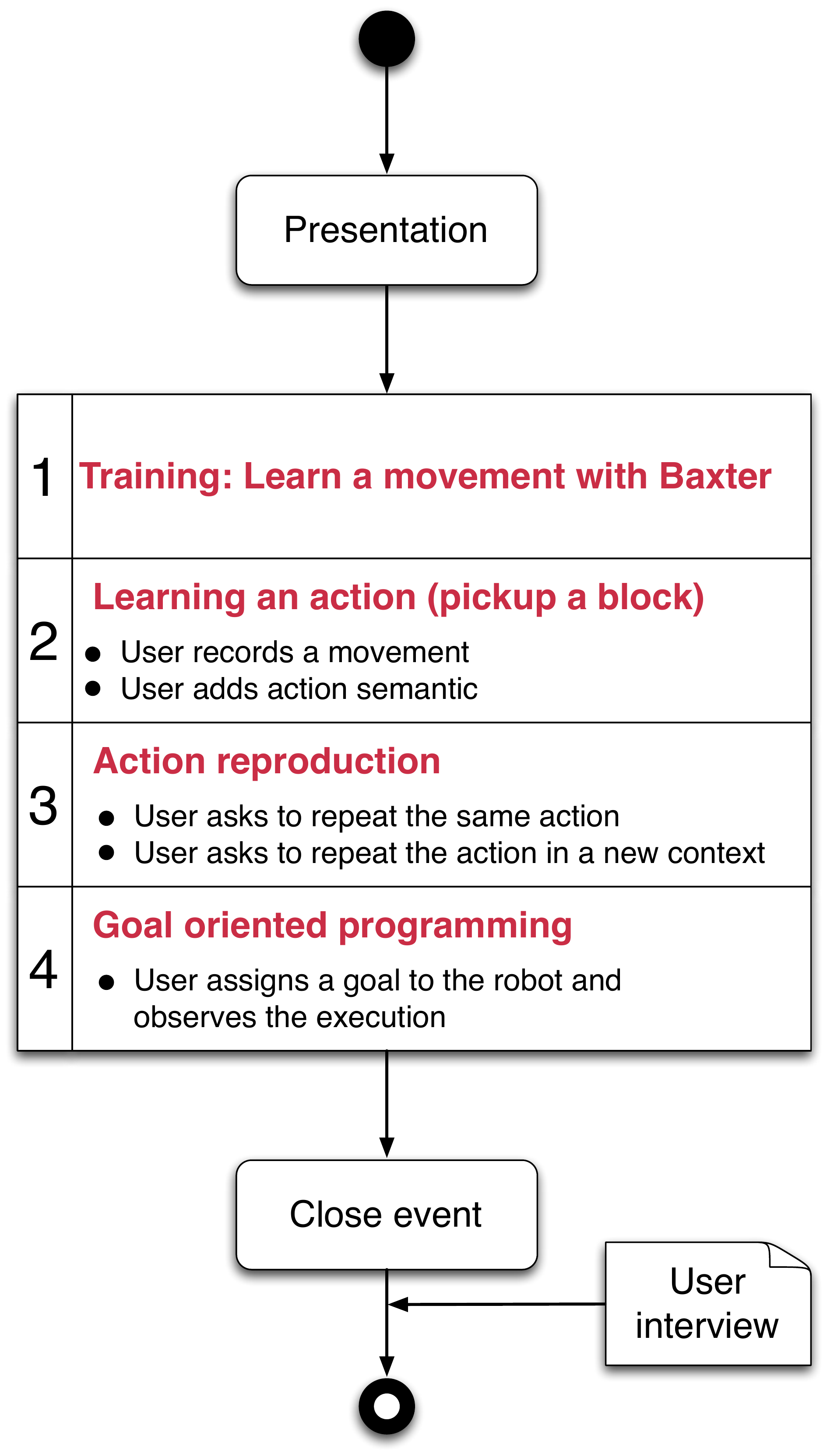}
    \caption{Overview of the experimental protocol}
    \label{fig:Experimental protocol}
\end{minipage}
\end{figure}
\subsection{Experimental Setup}
The experiments were conducted using a Baxter Research Robot from Rethink Robotics, with 7 DoF (per arm) and a maximum load of 2.27kg.
Both end effectors are mounted with a camera and an infrared sensor range. During the experiment the participants only manipulated Baxter's right limb which was equipped with a vacuum gripper. The robot's main head display was used to give visual feedback of the state of the system, e.g. the end effector's camera feed was constantly displayed when searching for objects, approval and refusal were expressed with the social feedback cues of nodding and shaking respectively.
Figure \ref{fig:Experimental set up} shows the experimental setup. The participants operated on a table with two positions marked with A and D (arrival and departure respectively), two square blocks (blue and red), that represent parts on an assembly line.

\subsection{Principal Hypotheses}\label{hypotheses}
The main goal of the experiment was to evaluate the usability of our proposed framework for non-experts. We stated the following principal hypotheses for our experiments:
\begin{enumerate}
\item \textit{A user without any programming knowledge is able to teach Baxter a moveObject action.} This allows us to evaluate the user's understanding of the basic PbD paradigm.
\item \textit{The user understands the concepts of action models, preconditions, and effects as used in Automated Planning techniques.} This allows us to verify whether the chosen logical representation for atomic action models in terms of preconditions and effects can be adopted easily by users without any programming background.
\item \textit{The user perceives Baxter as having learned a new task.} This relates to the user's overall satisfaction with the programming results.
\end{enumerate}

\subsection{Experimental Design}
\begin{figure}[h]
   \centering
    \includegraphics[scale=0.54]{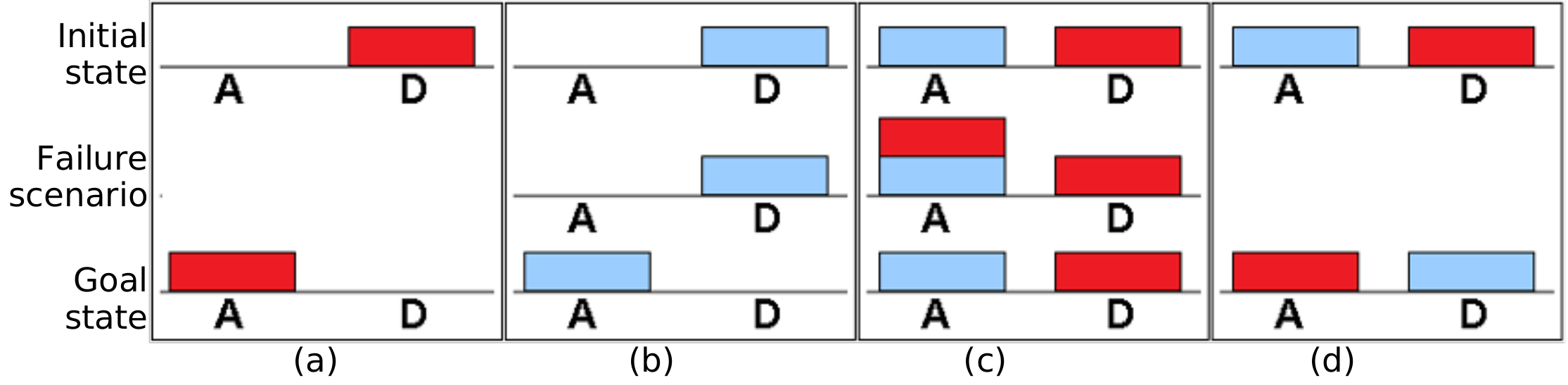}
    \caption{The four scenarios demonstrated to the participant showing the start (top) and the end (bottom) state after the execution of the latest taught action model. (b) and (c) show the failure scenarios before the action model has been modified by the user.}
    \label{fig:scenarios}
\end{figure}

To evaluate the user's understanding of preconditions and effects when creating new action models we consider three scenarios of incremental complexity.
The experiment takes place on a simulated assembly line, where objects of the same shape but different colour arrive consecutively at zone D (for departure).
The participant is asked to teach Baxter the action for moving an object from zone D to zone A (for arrival) by direct manipulation of Baxter's right limb. For each scenario the participant reviews the taught action model (in terms of preconditions and effects) and improves it. Starting with a simple naive action model, each scenario leads the participant closer towards a generalised representation of the moveObject action model. Figure \ref{fig:scenarios} illustrates the initial and final states of each scenario.

\noindent
The first scenario consists of the teaching phase using a single red block positioned on D. The participant demonstrates the moveObject action from D to A by direct manipulation of Baxter's limb. The demonstrated action is assigned:
\begin{itemize}
    \item[]{\tt moveObject:}
    \item preconditions: \textit{`red block on zone D'}
    \item effects: \textit{`red block on zone A'}
\end{itemize}
The level of difficulty is low to allow the participant to get acquainted with the logical representation. This is the only scenario where the user has to manipulate the robot's limb.

The second scenario uses a single blue block positioned on D. The participant is asked if the previously taught action model {\tt moveObject:} can be applied to the new block in order to move it from D to A. The participant is demonstrated a failure scenario (Figure \ref{fig:scenarios} (b)) when trying to apply the previous action model. We requested them to modify the conditions to evaluate their thinking on the logical representation which should result in:
\begin{itemize}
    \item[] {\tt moveObject:}
    \item preconditions: \textit{`block on zone D'}
    \item effects: \textit{`block on zone A'}
\end{itemize}
\noindent
The third scenario uses a blue block positioned on A and a red block positioned on D. The participant is asked if the latest action model can be applied to the new scenario in order to move the red block from D to A. The participant is demonstrated a scenario where the action model is incorrectly applied -- not considering the occupied arrival position (Figure \ref{fig:scenarios} (c)). The participant is requested to modify the action model to enable a correct moveObject behaviour. The additional condition demands a generalised thinking of higher complexity which should result in:
\begin{itemize}
    \item[] {\tt moveObject:}
    \item preconditions: \textit{`block on zone D', `zone A is empty'}
    \item effects: \textit{`block on zone A', `zone D is empty'}
\end{itemize}
\noindent
The forth scenario uses a blue block positioned on A and a red block positioned on D and a new position M. The participant is asked if the latest action model can be applied to a goal, namely to exchange the positions of the two blocks from D to A. The participant is demonstrated a scenario where the action model is used in combination with an automated planner to solve the permutation.

\subsection{Procedure}
The complete experimental protocol is shown in Figure \ref{fig:Experimental protocol}. Each experiment starts with an introduction to the Baxter robot followed by a training phase where users are given time to familiarise themselves with the manipulation of Baxter's limb.

The second phase consists of teaching Baxter to move a block from the departure to the arrival position. Users are requested to verify the extracted conditions of the state of the world and and raise any uncertainties.

The following phase consists of an iterative protocol to show the reproduction of the learned action: The user is presented a scenario and asked to describe their expectations when applying the latest action model. The user is faced with a failure scenario when trying to apply the action model in a new context and asked to present improvements to the existing action model to handle the new situations.

The final phase presents the user with a scenario where the learned action model is applied in combination with an automated planner to demonstrate a goal-oriented behaviour. The Baxter robot is assigned a goal and autonomously executes an action sequence to achieve the given goal.
Finally, the user is given a questionnaire to rate their experience and to provide us with their background information.

\section{Results}          
\label{S:Results}           
\begin{figure}[h]
    \centering
    \includegraphics[scale=0.8]{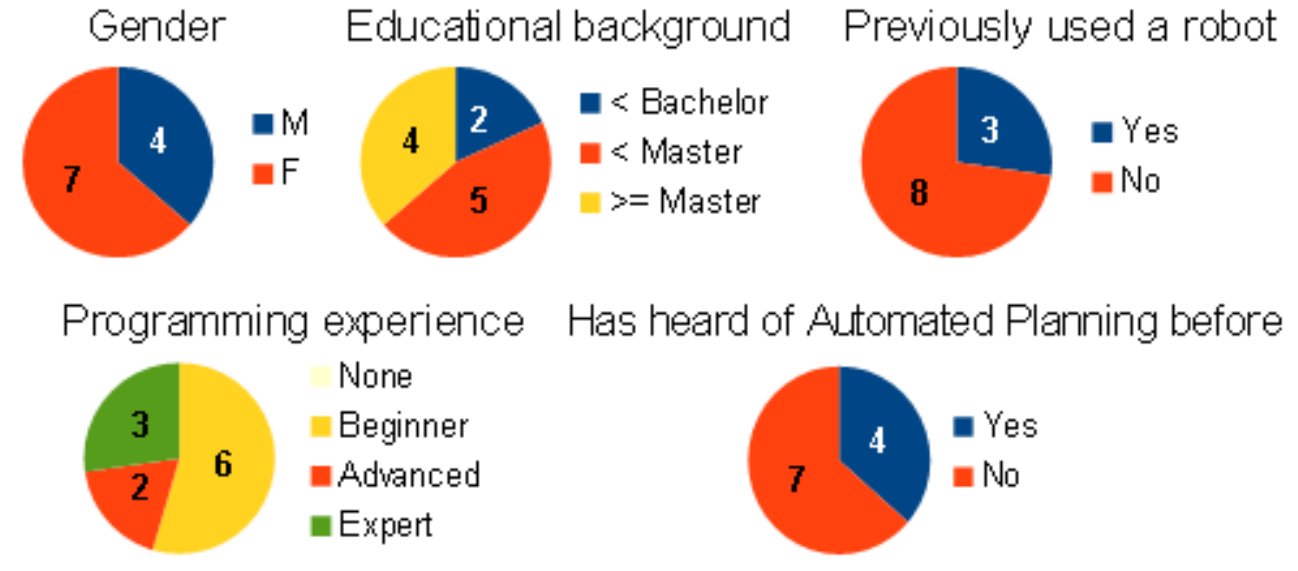}
   \caption{Overview of the participants recruited for the experiment}
    \label{fig:Participants}
  \end{figure}

We recruited 11 participants with diverse educational backgrounds, but focused on non-programming experts that had no knowledge of automated planning techniques.
Figure \ref{fig:Participants} shows the distribution of the participants for the experiments. The participants for `Programming experience' ranged from 0 `None' to 6 `Beginner' (experience in office productivity software), 2 `Advanced' (attended a programming course) to 3 `Expert' (studies in Computer Science). 3 of the participants who had previously heard of Automated Planning attended a related course.
Each participant was allocated 1 hour, with an overall average duration of 29 minutes and 32 seconds. The experiment was recorded while the participant interacts with the robot, until they are given the questionnaire at the end.
\noindent
Overall, all users were satisfied with the PbD process and Baxter's abilities to learn and reproduce the demonstrated moveObject action. All users understood and validated the extracted action model and managed to adopt the notion of preconditions and effects easily. As expected, no user managed to point out the missing conditions (i.e. generalisation on colours or consideration of empty zones) immediately at the start when asked for improvements. However, all users detected them easily, when faced with the relevant failure scenarios. The majority of users had difficulties with formulating the missing precondition (\textit{`zone A is empty'}) and suggested other equivalent conditions (\textit{`Do not place the object on zone A, if it is occupied'}). We conclude that the user interface should propose a pre-defined set of conditions that can be added to the action model.

Throughout the experiment some users made wide assumptions about the robot's capabilities. In the third scenario, where both arrival and departure zones were occupied (Figure \ref{fig:scenarios} (c)), almost half of the users expected Baxter to consider the occupied position, even though it was not mentioned in the Baxter's action model.
This is a common problem in PbD solutions as there is a difference in the perception of the robot's intelligence perceived by its teacher \cite{suay2012practical}. This can be easily addressed by reproducing the learned task in a new context and verifying the robot's knowledge base as we did throughout the experiment.

In the final phase users with no experience in automated planning did not expect Baxter to solve the permutation problem and agreed unanimously that it acted in an intelligent manner when it did. At the end of the experiment, all users stated that they had taught Baxter a new task and the majority understood the representation of the action models well. Finally, none of the users encountered any difficulties during the experiment.

\begin{figure}[h]
   \centering
    \includegraphics[scale=0.82]{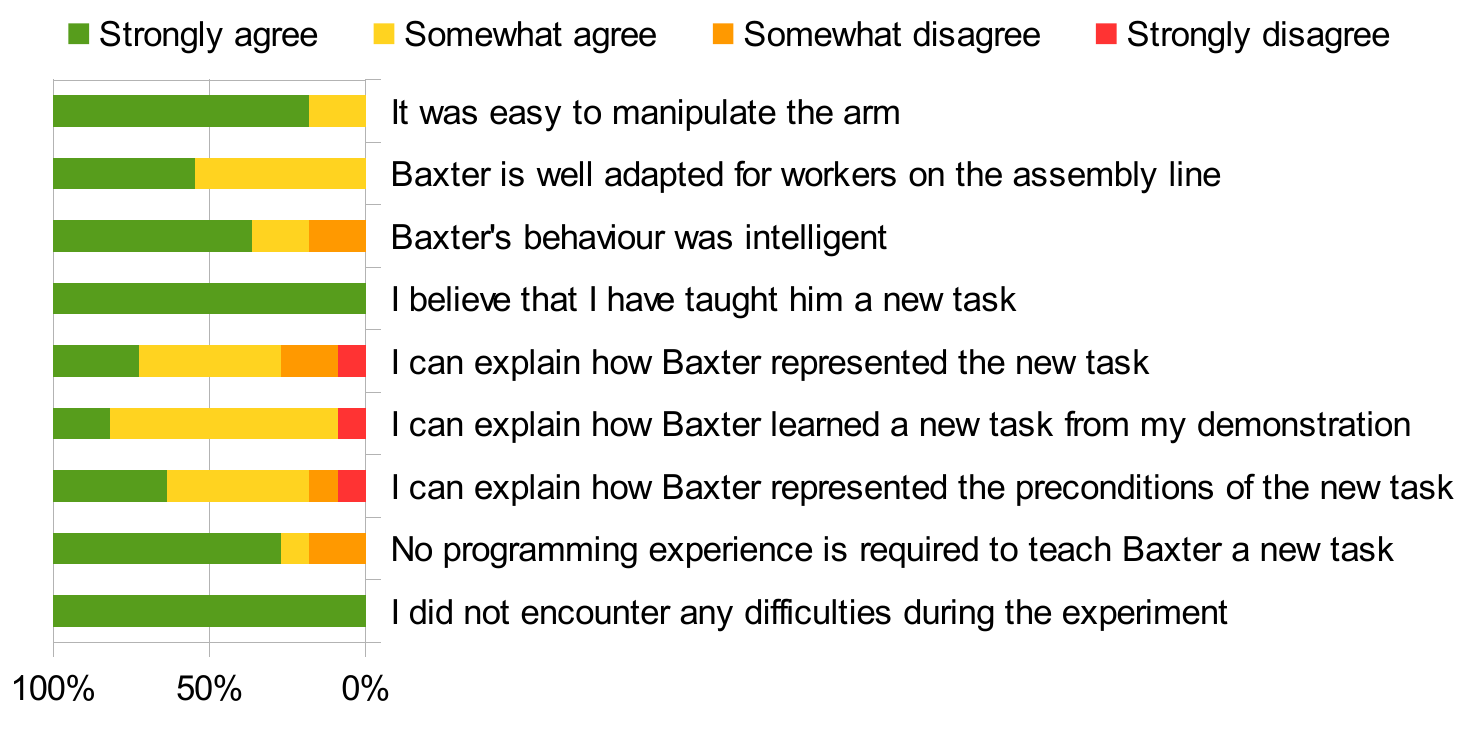}
    \caption{Responses to the questionnaire completed at the end of the experiment.}
    \label{fig:eEvaluation}
\end{figure}

With this experiment we verified the three principal hypotheses from Section \ref{hypotheses}. We showed that the level of representing an action in terms of preconditions and effects is adequate for non-expert users and that they can easily teach a robot a new action model to be used by an automated planner. Figure \ref{fig:eEvaluation} shows the responses from the questionnaire completed after the final phase.

\section{Conclusion and Future Work} 
\label{S:Conclusion}           

In this work we proposed a framework that allows non-expert users to teach a robot atomic actions with their semantic meanings that can be used by automated planners. Our framework combines two disciplines, Programming by Demonstration (PbD) and Automated Planning, and exploits the common logical representation in terms of preconditions and effects. We evaluated the user's understanding of the notion of these logical representations by conducting qualitative experiments.
Our experiments showed that users with and without programming experience understood the concept of PbD, as well as the notion of preconditions and effects, despite learning about them for the first time. Overall, the PbD process was considered to be very intuitive and easily understood by users. However, users were not able to construct a complete action model for the moveObject action without being faced with failure scenarios in contexts different to the initial demonstration. The complete vocabulary consisting of possible logical conditions should be proposed when creating the action model, as the exact logical formulation is not always straightforward.

After the final phase users believed to have taught Baxter a new moveObject action and considered its application to solve the permutation problem as intelligent. Overall our work demonstrated that the proposed framework with the logical representation of its action models is well-adapted to users with no programming experience.

Future work from an experimental point of view should focus on quantitative experiments to test the acceptability of a fully implemented system. We suggest the use of a richer domain including multiple scenarios and tasks taught in order to cover a wider set of problems. We will work towards in-situ experiments in an industrial environment to validate our found results. Higher human-robot interaction levels allowing multi-modal communication (vision, gesture, voice) may enhance the user experience, allowing them to easily clarify ambiguous situations (e.g. pointing at an object).
From a scientific point of view, further studies should focus on using statistical methods or efficient PbD solutions to generalise over demonstrated tasks.

\end{document}